\documentclass[letterpaper, 10 pt, conference]{ieeeconf}  

\IEEEoverridecommandlockouts                              
\usepackage[utf8]{inputenc}
\usepackage[T1]{fontenc}
\usepackage{graphicx}
\usepackage{float} 
\usepackage{ragged2e}
\usepackage{booktabs}
\usepackage{bbding}
\usepackage{multirow}
\usepackage{makecell}
\usepackage{amsmath,amssymb,amsfonts, bm}
\usepackage[table,xcdraw]{xcolor}
\usepackage{cite}
\usepackage[colorlinks,linkcolor=blue]{hyperref}
\usepackage{cleveref}
\usepackage{tabularx}
\usepackage[caption=true,font=normalsize,labelfont=sf,textfont=sf]{subfig}
\usepackage{textcomp}
\usepackage{stfloats}
\usepackage{tabularx}
\usepackage[linesnumbered,ruled,vlined]{algorithm2e}
\usepackage{adjustbox} 

\hypersetup{
    colorlinks=true, 
    linkcolor=blue,  
    filecolor=blue,  
    urlcolor=black,   
    citecolor=blue,  
    pdfborder={0 0 0} 
}

\Crefname{figure}{Fig.}{Figs.}
\Crefname{equation}{Eq.}{Eqs.}

\def\ie{\emph{i.e.}}

\title{\bf Statistic-Augmented, Decoupled MoE Routing and Aggregating \\ in Autonomous Driving
}

\author{~~~Wei-Bin Kou$^{1,2,3}$, Guangxu Zhu$^{3}$, Jingreng Lei$^{1}$, Chen Zhang$^{1}$, and Yik-Chung Wu$^{1,*}$, Jianping Wang$^{2,*}$
\thanks{$^{*}$Corresponding author: Yik-Chung Wu (ycwu@eee.hku.hk) and Jianping Wang (jianwang@cityu.edu.hk).}
\thanks{$^{1}$Department of Electrical and Electronic Engineering, The University of Hong Kong, Hong Kong 999077, China.}
\thanks{$^{2}$Department of Computer Science, City University of Hong Kong, Kong Hong 999077, China.}
\thanks{$^{3}$Shenzhen International Center For Industrial And Applied Mathematics, Shenzhen Research Institute of Big Data, Shenzhen, China.}
}

\begin{document}

\maketitle
\thispagestyle{empty}
\pagestyle{empty}

\begin{abstract}
Autonomous driving (AD) scenarios are inherently complex and diverse, posing significant challenges for a single deep learning model to effectively cover all possible conditions, such as varying weather, traffic densities, and road types. Large Model (LM)-Driven Mixture of Experts (MoE) paradigm offers a promising solution, where LM serves as the backbone to extract latent features while MoE serves as the downstream head to dynamically select and aggregate specialized experts to adapt to different scenarios. However, routing and aggregating in MoE face intrinsic challenges, including imprecise expert selection due to flawed routing strategy and inefficient expert aggregation leading to suboptimal prediction. To address these issues, we propose a statistic-augmented, decoupled \underline{MoE} \underline{R}outing and \underline{A}ggregating \underline{M}echanism (MoE-RAM) driven by LM. Specifically, on the one hand, MoE-RAM enhances expert routing by incorporating statistical retrieval mechanism to match LM-extracted latent features with cached prototypical features of the most relevant experts; on the other hand, MoE-RAM adaptively reweights experts' outputs in fusion by measuring  statistical distances of experts' instant features against LM-extracted latent features. Benefiting from the synergy of the statistic-augmented MoE's routing and aggregating, MoE-RAM ultimately improves the prediction performance. We take the AD semantic segmentation task as an example to assess the proposed MoE-RAM. Extensive experiments on AD datasets demonstrate the superiority of MoE-RAM compared to other MoE baselines and conventional single-model approaches.
\end{abstract}

\section{INTRODUCTION}
Autonomous driving (AD) stands at the forefront of artificial intelligence applications, demanding models that can process and respond to an extraordinarily diverse real-world scenarios in real time \cite{9981567,10160999,9811702,10416354,10049523}. These scenarios encompass a broad spectrum of complexities \cite{kou2024fedrc,kou2024fast,kou2025fedema}, including but not limited to dense urban traffic with unpredictable pedestrian movements, high-speed highway navigation under varying lighting conditions, rural roads with irregular terrains, and adverse weather conditions such as heavy rain, fog, or snow that obscure sensor inputs \cite{kou2025imacsr}. The inherent variability in these environments poses a fundamental limitation for a single deep learning model. Such models often overfit to dominant patterns in training data, failing to generalize to edge cases or unseen situations, which can result in critical safety risks in AD.

Large Model (LM)-Driven Mixture of Experts (MoE) \cite{mu2025comprehensive,kudugunta2021beyond,li2024boosting,mercurius2024amend} paradigm emerges as an promising solution to tackle these AD shortcomings. Specifically, a pretrained LM serves as the backbone to extract latent features of input, while multiple parallel fine-tuned "experts" (\ie, MoE) collectively serve as the downstream head to conduct relevant tasks, such as object detection, semantic segmentation, etc. In general, a routing (aggregating) mechanism dynamically selects (aggregates) the most relevant experts based on the input. 
However, the AD-oriented MoEs are hampered by their inherent challenges in routing and aggregating. First, existing MoE routing mechanisms typically rely on gating functions \cite{nguyen2024statistical,puigcerver2023sparse} (e.g., softmax-based functions) which are sensitive to feature mismatches. In dynamic AD environments, these routing functions often fail to accurately map inputs to the optimal experts, leading to suboptimal prediction accuracy. Second, inefficient aggregation compounds MoE's prediction performance, as conventional methods (e.g., simple weighted averaging) do not adequately capture inter-expert statistical dependencies, resulting in underestimated fusion activations. This inefficiency is particularly pronounced in non-IID (non-independent and identically distributed) AD scenarios.

To mitigate these challenges in MoE routing and aggregating and further consolidate MoE in AD application, we propose a novel LM-Driven \underline{MoE} framework with statistic-augmented, decoupled \underline{R}outing and \underline{A}ggregating \underline{M}echanism (MoE-RAM). Specifically, MoE-RAM comprises following integral components: (I) Pretrained ViT Backbone: we propose to use a pretrained Vision Transformer (ViT) as the backbone to extract latent features of each sensory image, which serve as the input of the downstream MoE architecture. (II) Downstream MoE with Expert-Wise Feature Retrieval Library (FRL): following the pretrained ViT backbone, we propose to deploy multiple CNN experts parallelly to form the MoE architecture to serve as the downstream task head. Each expert contains a FRL to store itself prototypical feature patterns, which represent individual expert's expertise and knowledge. FRL is updated by a read-then-update manner by using attention-based method. (III) Statistical Retrieval-Augmented MoE Routing Mechanism (MoE-RM): we enhance the MoE routing process by integrating a retrieval mechanism that statistically queries each expert's FRL to identify the most relevant experts. This is achieved through measuring Jensen-Shannon divergence to align ViT-extracted latent features with each expert expertise in itself FRL, enabling precise, scenario-aware expert selection. For example, in a foggy driving scenario, MoE-RM retrieves experts who specialize in low-visibility driving scenario's prediction, ensuring adaptive and accurate prediction. (IV) Statistical Similarity-Augmented MoE Aggregating Mechanism (MoE-AM): to overcome the flawed experts' fusion policy, we propose a statistical similarity-based expert aggregation mechanism that employs dynamic statistical reweighting scheme, which aggregates expert outputs by considering their statistical distribution alignments. Specifically, we firstly calculate statistical distances of all experts' instant features against ViT-extracted latent features using Jensen-Shannon divergence; we then assign aggregation weight for each expert by computing the proportion of its distance reciprocal relative to the total sum of distance reciprocal of all experts.

\begin{figure*}[tp]
\hspace{-0.6cm}
\includegraphics[width=1.035\linewidth]{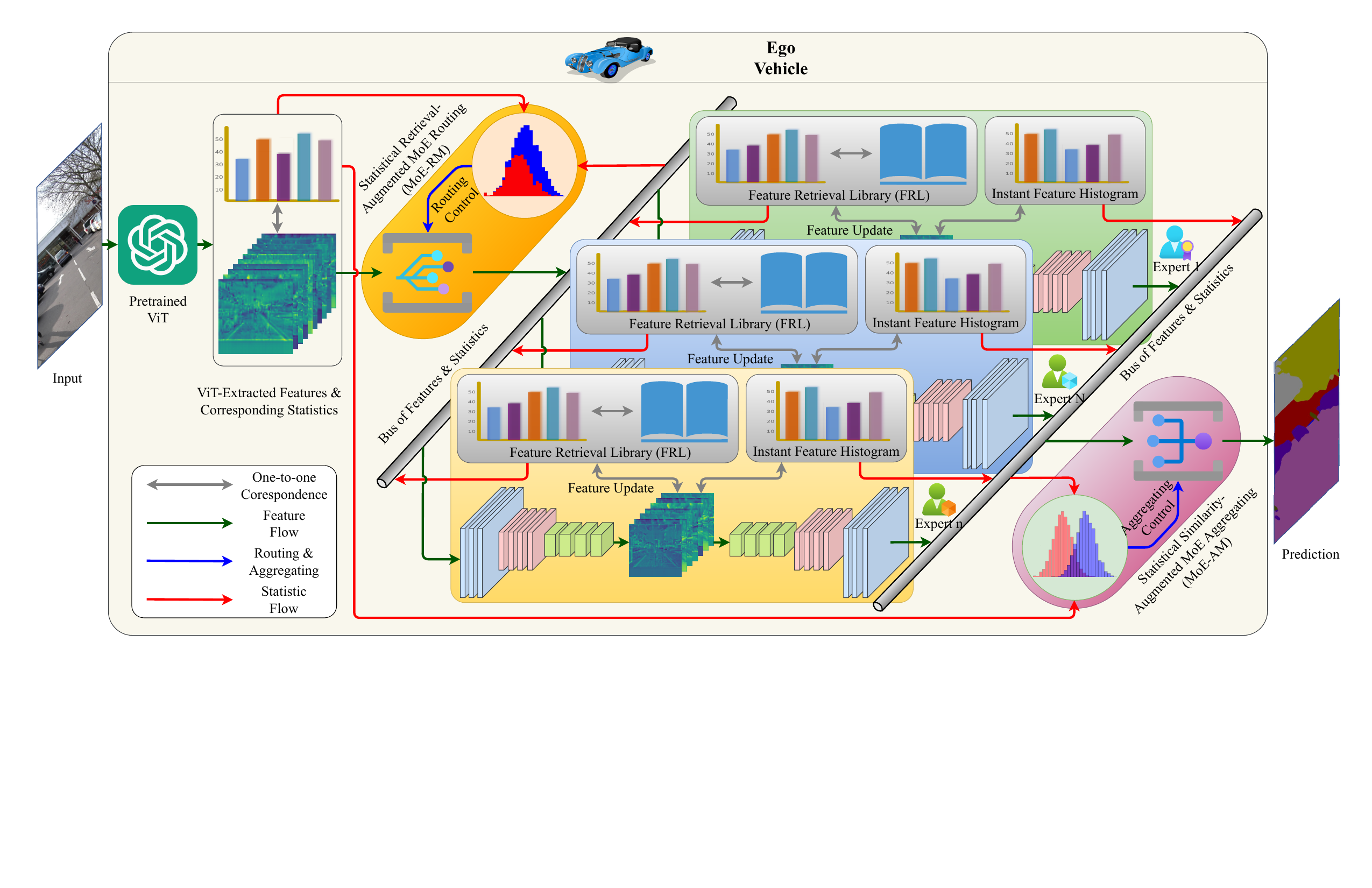}
\vspace{-4.1cm}
\caption{Illustration of the proposed MoE-RAM.}
\label{fig:MoE-RAM-overview}
\vspace{-0.7cm}
\end{figure*}

By synergistically integrating abovementioned routing and aggregating mechanisms (\ie, MoE-RM and MoE-AM), the proposed MoE-RAM transforms MoE into a more robust and intelligent system, tailored to the AD demands where precision and adaptability are paramount. The proposed MoE-RAM is illustrated in \Cref{fig:MoE-RAM-overview}. We take the AD semantic segmentation task as an example to evaluate the proposed MoE-RAM. Extensive experiments on AD benchmarks demonstrate that the proposed MoE-RAM outperforms other MoE baselines and single-model competitors.

Main contributions of this work are highlighted as follows:
\begin{itemize}
    \item MoE routing and aggregating generally impair the prediction accuracy for AD-oriented MoE systems. To mitigate this issue, we propose MoE-RAM to enhance the routing and aggregating of MoE from statistic perspective. The proposed MoE-RAM comprises a pretrained ViT backbone, expert-wise FRLs, a statistical retrieval-augmented MoE routing mechanism (\ie, MoE-RM), and a statistical similarity-augmented MoE aggregating mechanism (\ie, MoE-AM).
    \item Expert-wise FRLs are updated in a read-then-update fashion based on attention mechanism to store experts' expertise and knowledge. MoE-RM deploys both the retrieved experts' statistic from respective FRL and ViT-extracted features' statistic to determine which subset of experts is selected. MoE-AM dynamically assigns weights for involved experts in MoE aggregation, which is achieved by measuring statistical alignment of participated experts' features with the ViT-extracted features. 
    \item Extensive experiments on AD datasets show that the proposed MoE-RAM outperforms other MoE baselines and single-model competitors in prediction accuracy.
\end{itemize}

\section{Related Work}
\label{related_work}
\subsection{Foundation Models Towards AD}
Recent advancements in Large Language Models (LLMs), such as the GPT \cite{achiam2023gpt} and LLaMA \cite{touvron2023llama}, have revolutionized natural language processing. Similarly, models such as LLaVA \cite{liu2024llavanext}, ShareGPT4V \cite{chen2023sharegpt4v} integrate visual information to enable tasks like image captioning and visual question answering. These Vision-Language Models (VLMs) are typically trained on extensive image-text datasets \cite{ALIGN}. Early VLMs focused on static images \cite{CLIP}, but recent efforts have extended their capabilities to video understanding such as Qwen2-VL~\cite{wang2024qwen2}, incorporating temporal dynamics into the language feature space for sequence comprehension. Several recent works explore the potential of VLMs for AD \cite{kou2024pfedlvmlargevisionmodel}. For example, Dolphins~\cite{ma2023dolphins} employs Visual Question Answering (VQA) to bridge the gap between data-driven driving and user trust. Besides, decision-making and planning are also being integrated into VLMs, as seen in DriveVLM~\cite{tian2024drivevlm} and Reason2drive~\cite{nie2025reason2drive}, where the training data is always divided into perception, prediction, and planning components. These models often produce outputs, including scene descriptions, action analysis, hierarchical planning, etc. End-to-end AD approaches such as DriveGPT4~\cite{xu2024drivegpt4} and LMDrive~\cite{shao2024lmdrive} attempt to directly map visual and linguistic inputs to planning or low-level control signals. 

\subsection{Mixture of Experts (MoE) Towards AD}
The MoE architecture has emerged as a pivotal method for scaling LMs and enhancing task specialization. By employing MoE designs, model capacity and processing efficiency are significantly improved \cite{mu2025comprehensive}. The primary strength of MoE lies in its ability to harness the specialization of individual experts across diverse data, thereby boosting overall model performance. For instance, \cite{kudugunta2021beyond} applies a task-level MoE to multilingual translation, routing inputs intelligently based on task identifiers, resulting in notable performance gains and improved inference efficiency. While MoE has shown great promise in LLMs and VLMs, its application in AD remains relatively underexplored. Existing studies have examined MoE architectures in AD for specific tasks, such as rare scenario perception \cite{li2024boosting}, long-tailed trajectory prediction \cite{mercurius2024amend}, domain adaptation under varying weather conditions \cite{kim2022learning}, safe trajectory prediction and planning \cite{pini2022safe}, and generalization of planners \cite{sun2024generalizing}. However, these approaches have yet to fully exploit MoE's potential to deliver adaptive and robust performance across diverse and complex AD scenarios.

\subsection{MoE's Routing and Aggregating}
MoE routing is generally achieved through a gating function that assigns inputs to preferred experts based on specific criteria. Linear gating functions \cite{riquelme2021scaling,fan2022m3vit}, such as softmax-based gating, are commonly used due to their simplicity and effectiveness. These functions compute a probability distribution over experts and often include a TopK operation to select the top-k experts. Non-linear alternatives \cite{li2022sparse,nguyen2024statistical} such as cosine similarity-based gating have been introduced to enhance generalization. These approaches generally project inputs into a hyperspherical space and compare them to expert embeddings. SoftMoE \cite{puigcerver2023sparse} explore exponential family distributions for weighted averaging to avoid discrete token dropping.
MoE aggregation combines the outputs of selected experts by weighted summation, where weights are determined by the gating function. For example, the output of Transformer-based MoE is calculated as a linear combination of the outputs of feed-forward networks (FFNs) of all selected experts  \cite{riquelme2021scaling}. Variants such as mixture-of-attention (MoA) \cite{wang2024moa} dynamically aggregate attention heads. These aggregation mechanisms ensure efficient knowledge integration while addressing challenges such as load imbalance through auxiliary loss functions. In this paper, we consider the MoE's routing and aggregating separately from statistical perspective, and decouple the inter-dependancy of MoE's routing and aggregating, aiming at enhancing MoE's overall prediction performance.

\setlength{\textfloatsep}{1pt}
\begin{table}[t]
\centering
\caption{Key Notations in MoE-RAM}
\label{tab:notations_moe_ram}
\setlength{\tabcolsep}{3pt}
\renewcommand{\arraystretch}{1.1}
\begin{tabularx}{\linewidth}{ll}
\hline
Notation & Explanation \\
\hline
$\mathcal{D}$ & Dataset of RGB images. \\
$\mathcal{D}^{(i)}$ & The $i$-th input RGB image of $\mathcal{D}$. \\
$\bm{\omega}_{\mathrm{ViT}}$ & Parameters of the pretrained ViT backbone. \\
$\mathcal{F}^{(i)}_{\mathrm{ViT}}$ & ViT-extracted latent features of $\mathcal{D}^{(i)}$. \\
$N$ & Number of experts in MoE. \\
$\mathcal{E}_j, \bm{\theta}_j$ & The $j$-th CNN expert and its parameters. \\
$\mathcal{E}^{\mathrm{enc}}_j, \bm{\theta}^{\mathrm{enc}}_j$ & Encoder of expert $j$ and its parameters (former part of $\mathcal{E}_j$). \\
$\mathbf{h}^{(i)}_j$ & Intermediate representations of expert $j$ for input $i$. \\
$\mathbf{y}^{(i)}_j$ & Task-specific output of expert $j$ for input $i$. \\
$\mathcal{M}_j$ & Feature Retrieval Library (FRL) of expert $j$. \\
$K_j$ & Number of prototype entries in $\mathcal{M}_j$. \\
$\mathbf{p}_{j,k} \in \mathbb{R}^{d}$ & The $k$-th prototype vector in expert $j$’s FRL. \\
$w_{j,k} \ge 0$ & Importance weight associated with prototype $\mathbf{p}_{j,k}$. \\
$\alpha^{(i)}_{j,k}$ & Attention weight on prototype $\mathbf{p}_{j,k}$ for input $i$. \\
$\widetilde{\mathbf{p}}^{(i)}_j$ & Retrieved prototype from expert $j$ for input $i$. \\
$\eta$ & FRL memory update rate for read-then-update. \\
$Q^{(i)}$ & Input-induced normalized distribution from $\mathcal{F}^{(i)}_{\mathrm{ViT}}$. \\
$P^{(i)}_j$ & Prototype-induced normalized distribution from $\widetilde{\mathbf{p}}^{(i)}_j$. \\
$R^{(i)}_j$ & Output-induced normalized distribution from $\mathbf{h}^{(i)}_j$. \\
$s^{(i)}_j$ & Routing score for expert $j$ on input $i$. \\
$\pi^{(i)}_j$ & Routing probability for expert $j$ on input $i$. \\
$\mathcal{S}^{(i)}$ & Index set of TopK selected experts for input $i$. \\
$\delta^{(i)}_j$ & JS divergence between $Q^{(i)}$ and $R^{(i)}_j$. \\
$\widetilde{w}^{(i)}_j$ & Reciprocal distance weight $1/(\epsilon+\delta^{(i)}_j)$ for aggregation. \\
$\beta^{(i)}_j$ & Aggregation weight for expert $j$, normalized over $j\in\mathcal{S}^{(i)}$. \\
$\widehat{\mathbf{y}}^{(i)}$ & Final aggregated output for input $i$. \\
$B$ & Mini-batch size. \\
\hline
\end{tabularx}
\end{table}

\section{Methodology}
\label{methodology}
The proposed MoE-RAM is composed of a pretrained ViT backbone, expert-wise FRLs, a statistical retrieval-augmented routing mechanism  MoE-RM, and a statistical similarity-augmented aggregating mechanism MoE-AM. They are introduced in \Cref{method_A}, \Cref{method_B}, \Cref{method_C}, and \Cref{method_D}, respectively. Finally, we detail the overall training objective of MoE-RAM in \Cref{method_E}.

\subsection{Pretrained ViT Backbone}
\label{method_A}
Pretrained ViT \cite{dosovitskiy2020image} is a widely used representation extractor to extract latent features of RGB data. Therefore, in this work, we, as usual, select ViT as the LM backbone (with parameters $\pmb{\omega}_{ViT}$) to extract the latent features of inputs. 

We denote the $i$-th input RGB image as $\mathcal{D}^{(i)}$ for dataset $\mathcal{D}$, and its latent features $\mathcal{F}_{ViT}^{(i)}$ is extracted in a zero-shot manner as follow 
\begin{align}
\mathcal{F}^{(i)}_{ViT} = \pmb{\omega}_{ViT}(\mathcal{D}^{(i)}),
\label{Eq:ViT_seed_forward}
\end{align}
where $\mathcal{F}^{(i)}_{ViT}$ is a high-dimensional tensor, representing the ViT's understanding of the content of input $\mathcal{D}^{(i)}$. 

\subsection{Expert-Wise Feature Retrieval Library (FRL)}
\label{method_B}
Following the pretrained ViT backbone, we deploy a downstream sparse MoE head composed of $N$ parallel CNN experts $\{\mathcal{E}_j\}_{j=1}^{N}$, each parameterized by $\pmb{\theta}_j$. To explicitly encode and preserve each expert’s specialization and expertise, we equip every expert $\mathcal{E}_j$ with a Feature Retrieval Library (FRL) denoted as $\mathcal{M}_j$. The FRL stores prototypical feature patterns that summarize the expert’s learned expertise and scenario-specific knowledge.

Given the ViT-extracted latent feature $\mathcal{F}^{(i)}_{ViT}$ from Eq.~\eqref{Eq:ViT_seed_forward}, expert $\mathcal{E}_j$ produces an intermediate representation $\mathbf{h}^{(i)}_j$ and a task-specific output $\mathbf{y}^{(i)}_j$:
\begin{align}
\mathbf{h}^{(i)}_j &= \mathcal{E}_j^{\text{enc}}(\mathcal{F}^{(i)}_{ViT}; \pmb{\theta}^{\text{enc}}_j), 
\label{eq:inter_representation}
\\
\mathbf{y}^{(i)}_j &= \mathcal{E}_j(\mathcal{F}^{(i)}_{ViT}; \pmb{\theta}_j),
\label{eq:task_output}
\end{align}
where $\mathcal{E}_j^{\text{enc}}$ is the expert-embedded encoder producing an expert-level representation $\mathbf{h}^{(i)}_j$ for updating FRL, and noting that $\mathcal{E}_j^{\text{enc}}$ is the former part of $\mathcal{E}_j$ (\ie, $\pmb{\theta}_j^{\text{enc}} \in \pmb{\theta}_j$).

With respect to FRL of expert $\mathcal{E}_j$, $\mathcal{M}_j$ is a set of $K_j$ prototypical feature entries $\{(\mathbf{p}_{j,k}, w_{j,k})\}_{k=1}^{K_j}$ with vectors $\mathbf{p}_{j,k}\in\mathbb{R}^{d}$ and associated nonnegative importance weights $w_{j,k}$. We firstly compute attention weights $\alpha_{j,k}^{(i)}$ over $\mathcal{M}_j$ to read expert prototypes $\mathbf{p}_{j,k}$ relevant to $\mathcal{F}^{(i)}_{ViT}$:
\begin{align}
\alpha_{j,k}^{(i)} &= \frac{\exp(\mathrm{sim}(\mathcal{F}^{(i)}_{ViT}, \mathbf{p}_{j,k}))}{\sum\nolimits_{\ell=1}^{K_j}\exp(\mathrm{sim}(\mathcal{F}^{(i)}_{ViT}, \mathbf{p}_{j,\ell}))}, 
\label{eq:attention_weight}
\\
\widetilde{\mathbf{p}}^{(i)}_j &= \sum\nolimits_{k=1}^{K_j} \alpha_{j,k}^{(i)} \mathbf{p}_{j,k},
\label{eq:retrieved_proto}
\end{align}
where $\mathrm{sim}(\cdot,\cdot)$ is a similarity function (e.g., cosine similarity), $\widetilde{\mathbf{p}}^{(i)}_j$ is the retrieved prototype of expert $\mathcal{E}_j$ for input $\mathcal{D}_{(i)}$, serving the role of expert selection in MoE routing. 

Parallel with the forward pass, we perform a read-then-update step to adaptively refine $\mathcal{M}_j$ as follows:
\begin{align}
\mathbf{p}_{j,k} &\leftarrow (1-\eta \alpha_{j,k}^{(i)}) \mathbf{p}_{j,k} + \eta \alpha_{j,k}^{(i)} \psi_j(\mathbf{h}^{(i)}_j), 
\label{eq:update_FRL}
\\
w_{j,k} &\leftarrow (1-\eta \alpha_{j,k}^{(i)}) w_{j,k} + \eta \alpha_{j,k}^{(i)},
\end{align}
where $\eta$ is a memory update rate and $\psi_j(\cdot)$ maps the experts' intermediate features into the prototype space. This attention-weighted update maintains compact, scenario-aware prototypes that reflect the expert’s evolving competence.

\subsection{Statistic-Augmented MoE Routing (MoE-RM)}
\label{method_C}
To achieve precise, scenario-aware expert selection, we augment routing with statistical retrieval over FRLs. For expert $\mathcal{E}_j$, we estimate a prototype-induced distribution $P_j^{(i)}$ from $\widetilde{\mathbf{p}}^{(i)}_j$, and then align it to the input-induced distribution $Q^{(i)}$ from $\mathcal{F}^{(i)}_{ViT}$ using Jensen–Shannon (JS) divergence:
\begin{align}
Q^{(i)} &= \mathrm{Norm}(\mathcal{F}^{(i)}_{ViT}), \\
P_j^{(i)} &= \mathrm{Norm}(\widetilde{\mathbf{p}}^{(i)}_j), \label{eq:norm_pji} \\
\hspace{-0.3cm}D_{\mathrm{JS}}(Q^{(i)}||P_j^{(i)}) \!\!&=\!\! \tfrac{1}{2} (D_{\mathrm{KL}}(Q^{(i)} || I_j^{(i)}) \!\!+\!\! D_{\mathrm{KL}}(P_j^{(i)} || I_j^{(i)})), 
\end{align}
where $\mathrm{Norm}(\cdot)$ denotes a feature-to-distribution normalization (e.g., softmax over channels or bins), $D_{\mathrm{KL}}(\cdot||\cdot)$ is the Kullback–Leibler divergence, and $I_j^{(i)} = \tfrac{1}{2}(Q^{(i)} + P_j^{(i)})$.

We convert JS divergences into routing scores and probabilities based on following equations:
\begin{align}
s^{(i)}_j &= 1/(\epsilon + D_{\mathrm{JS}}(Q^{(i)}||P_j^{(i)})), \\
\pi^{(i)}_j &= \exp(\tau s^{(i)}_j)/(\sum\nolimits_{\ell=1}^{N}\exp(\tau s^{(i)}_\ell)),
\label{eq:routing_prob}
\end{align}
with small $\epsilon>0$ for numerical stability and temperature $\tau>0$ to control sharpness. Based on above scores, we apply TopK routing to $\{\pi^{(i)}_j\}_{j=1}^{N}$. The routed set $\mathcal{S}^{(i)}\subseteq \{1,\dots,N\}$ contains the indices of selected experts for input $\mathcal{D}^{(i)}$.

Intuitively, experts whose FRLs better match the current scenario (lower JS divergence) receive higher routing probabilities. For instance, under foggy scenes, experts with FRLs encoding low-visibility patterns are prioritized by MoE-RM.

\subsection{Statistic-Augmented MoE Aggregating (MoE-AM)}
\label{method_D}
To address suboptimal fusion of expert outputs, we propose a statistical similarity-based aggregation that dynamically reweights experts in fusion according to expert $\mathcal{E}_j$'s intermediate representation $\mathbf{h}^{(i)}_j$'s distribution alignment with that of $\mathcal{F}^{(i)}_{ViT}$.

Given per-expert intermediate representation $\{\mathbf{h}^{(i)}_j\}_{j=1}^N$ for ${j\in \mathcal{S}^{(i)}}$, we firstly compute intermediate representation-induced distributions $R^{(i)}_j$ over $\mathbf{h}^{(i)}_j$ and measure their JS divergence to input-induced distribution $Q^{(i)}$ over $\mathcal{F}^{(i)}_{ViT}$:
\begin{align}
R^{(i)}_j&=\mathrm{Norm}(\mathbf{h}^{(i)}_j),  \quad j \in \mathcal{S}^{(i)},
\label{eq:output_dist}
\\
\delta^{(i)}_j &= D_{\mathrm{JS}}(Q^{(i)}|| R^{(i)}_j), \quad j \in \mathcal{S}^{(i)}.
\label{eq:output_JS}
\end{align}
We subsequently define a reciprocal-distance weight and normalize across the routed experts as below equations:
\begin{align}
\widetilde{w}^{(i)}_j &= 1 / (\epsilon + \delta^{(i)}_j), \quad j \in \mathcal{S}^{(i)}, 
\label{eq:reciprocal_distance}
\\
\beta^{(i)}_j &= \widetilde{w}^{(i)}_j/(\sum\nolimits_{\ell\in \mathcal{S}^{(i)}} \widetilde{w}^{(i)}_\ell), \quad j \in \mathcal{S}^{(i)}.
\label{eq:norm_reciprocal_distance}
\end{align}
Finally, we aggregate outputs of routed experts with above statistical reweighted coefficient, \ie,
\begin{align}
\widehat{\mathbf{y}}^{(i)} &= \sum\nolimits_{j\in \mathcal{S}^{(i)}} \beta^{(i)}_j \mathbf{y}^{(i)}_j.
\label{eq:aggregated_output}
\end{align}
This aggregation emphasizes experts whose outputs are distributionally consistent with the ViT-extracted latent features, providing adaptive, robust fusion under diverse, complex AD scenarios.

\subsection{MoE-RAM's Overall Training Objective}
\label{method_E}
MoE-RAM can be trained in an end-to-end manner. In this work, we take the AD semantic segmentation task as an example to evaluate the proposed MoE-RAM. The overall training objective comprises three parts: the standard cross entropy loss $\mathcal{L}_{\text{CE}}$, the MoE auxiliary load-balancing term $\mathcal{L}_{\text{LB}}$, and FRL regularizer $\mathcal{L}_{\text{FRL}}$. Thus, the overall training objective $\mathcal{L}$ can be formulated as 
\begin{align}
\mathcal{L} = \mathcal{L}_{\text{CE}} + \lambda_{\text{LB}} \mathcal{L}_{\text{LB}} + \lambda_{\text{FRL}} \mathcal{L}_{\text{FRL}},
\label{eq:overall_obj}
\end{align}
where $\lambda_{\text{LB}}, \lambda_{\text{FRL}}\ge 0$ control regularization strengths. In the proposed MoE-RAM, term $\mathcal{L}_{\text{LB}}$ is introduced to avoid expert collapse in training by pushing experts routed uniformly, and is defined as  
\begin{align}
\mathcal{L}_{\text{LB}} = \sum\nolimits_{j=1}^{N} u_j \log u_j - \log N,
\label{eq:LB_obj}
\end{align}
where $u_j = \frac{1}{B} \sum\nolimits_{i=1}^{B} \pi^{(i)}_j, j=1,\dots,N$. $\sum\nolimits_{j=1}^{N} u_j \log u_j$ (minimized when $\{u_j\}_{j=1}^N$ is uniform) pushes to distribute traffic across all experts, avoiding expert collapse. Adding $-log N$ just zero-centers at the optimum. For the FRL regularizer $\mathcal{L}_{\text{FRL}}$, we propose to formulate it as 
\begin{align}
\mathcal{L}_{\text{FRL}} \!\!=&\!\!
\sum\nolimits_{j=1}^{N} \sum\nolimits_{k=1}^{K_j} (\underbrace{\lVert \mathbf{p}_{j,k} \rVert_2^2 \!+\! w_{j,k}^2}_{\text{Norm/Weight Decay}} \!+\!  \underbrace{\sum\nolimits_{i=1}^{B}| \alpha^{(i)}_{j,k}|}_{\text{Sparse Attention}}), 
\label{eq:FRL_obj}
\end{align}
where \textit{Norm/Weight Decay} prevents unbounded growth of prototype vectors and their importance scalars, controlling scale and avoiding trivial wins by increasing magnitude, and \textit{Sparse Attention} over prototypes encourages the retriever to pick a few relevant prototypes rather than averaging many, which keeps retrieved signals sharp and interpretable.

In conclusion, the proposed MoE-RAM is outlined in Algorithm \ref{alg:moe_ram}.

\setlength{\textfloatsep}{1pt}
\begin{algorithm}[tp]
\SetKwInput{KwInit}{Init}
\SetKwInput{KwIn}{Input}
\SetKwInput{KwOut}{Output}
\SetKwComment{tcp}{// }{}
\caption{MoE-RAM}
\label{alg:moe_ram}

\KwIn{Dataset $\mathcal{D}$; pretrained $\bm{\omega}_{\mathrm{ViT}}$; experts $\{\mathcal{E}_j\}_{j=1}^N$ with params $\{\bm{\theta}_j\}_{j=1}^N$; FRLs $\{\mathcal{M}_j\}_{j=1}^N$ with $\{(\mathbf{p}_{j,k}, w_{j,k})\}_{k=1}^{K_j}$; update rate $\eta$; temperature $\tau$; stability $\epsilon$; loss weights $\lambda_{\mathrm{LB}}, \lambda_{\mathrm{FRL}}$.}
\KwOut{Trained $\{\bm{\theta}_j\}_{j=1}^N$, and updated $\{\mathcal{M}_j\}_{j=1}^N$.}

\KwInit{
Initialize experts $\{\bm{\theta}_j\}_{j=1}^N$; initialize $\{\mathcal{M}_j\}_{j=1}^N$ with $\mathbf{p}_{j,k}\in\mathbb{R}^d$ and weights $w_{j,k}\ge 0$. \\
}

\For{each training step}{
  Sample a mini-batch $\{\mathcal{D}^{(i)}\}_{i=1}^{B}$ with labels; \\

  \tcp{1) Backbone feature extraction}
  \For{$i \leftarrow 1$ \KwTo $B$}{
    $\mathcal{F}^{(i)}_{\mathrm{ViT}} \leftarrow \bm{\omega}_{\mathrm{ViT}}(\mathcal{D}^{(i)})$, 
    $Q^{(i)} \leftarrow \mathrm{Norm}(\mathcal{F}^{(i)}_{\mathrm{ViT}})$
  }

  \tcp{2) FRL retrieval per expert}
  \For{$j \leftarrow 1$ \KwTo $N$}{
    \For{$i \leftarrow 1$ \KwTo $B$}{
      $\alpha^{(i)}_{j,k}, \widetilde{\mathbf{p}}^{(i)}_j, P^{(i)}_j \leftarrow $ \Cref{eq:attention_weight,eq:retrieved_proto,eq:norm_pji}
    }
  }

  \tcp{3) Expert Routing}
  \For{$i \leftarrow 1$ \KwTo $B$}{
    $\pi^{(i)}_j \leftarrow$ \Cref{eq:routing_prob},
    $\mathcal{S}^{(i)} \leftarrow \mathrm{Top}\mbox{-}K(\{\pi^{(i)}_j\}_{j=1}^N)$
  }

  \tcp{4) Expert forward \& aggregation}
  \For{$i \leftarrow 1$ \KwTo $B$}{
    \ForEach{$j \in \mathcal{S}^{(i)}$}{
      $\mathbf{h}^{(i)}_j, \mathbf{y}^{(i)}_j, R^{(i)}_j, \delta^{(i)}_j \leftarrow $ \Cref{eq:inter_representation,eq:task_output,eq:output_dist,eq:output_JS}
    }
    $\widetilde{w}^{(i)}_j, \beta^{(i)}_j, \widehat{\mathbf{y}}^{(i)} \leftarrow$ \Cref{eq:reciprocal_distance,eq:norm_reciprocal_distance,eq:aggregated_output}
  }

  \tcp{5) Optimization Losses}
  $\mathcal{L}_{\mathrm{LB}}, \mathcal{L}_{\mathrm{FRL}}, \mathcal{L} \leftarrow$ \Cref{eq:LB_obj,eq:FRL_obj,eq:overall_obj}

  \tcp{6) FRL read-then-update}
  \For{$j \leftarrow 1$ \KwTo $N$}{
    \For{$k \leftarrow 1$ \KwTo $K_j$}{
      $\mathbf{p}_{j,k} \leftarrow $ \Cref{eq:update_FRL}
    }
  }

  \tcp{7) Parameter update}
  Update $\{\pmb{\theta}_j\}_{j=1}^N$ with an optimizer step on $\mathcal{L}$.
}
\end{algorithm}

\vspace{-0.25cm}
\section{Experiments}
\label{experiments}
In this section, we carry out comprehensive experiments to verify MoE-RAM on AD semantic segmentation task. Specifically, we firstly introduce the experimental setup in \Cref{sec:exp_setup}. Subsequently, we conduct experimental comparison and related analyses in \Cref{sec:exp_comp_analyses}. We then visualize the relationship between ViT-extracted features and expert-wise FRL prototypes in \Cref{sec:MoERAM_visualization}. Finally, we conduct ablation studies to investigate various hyperparameters' effect on MoE-RAM's overall performance in \Cref{sec:ablation_study}.

\subsection{Datasets, Evaluation Metrics and Implementation} \label{sec:exp_setup}
\subsubsection{Datasets}
The \textbf{Cityscapes} dataset \cite{Cordts2016Cityscapes} includes 2,975 training and 500 validation images annotated with masks for 19 semantic classes, such as vehicles and pedestrians. The \textbf{CamVid} dataset \cite{brostow2008segmentation} contains 701 images across 11 semantic classes, with 600 used for training and 101 for testing. A subset of the \textbf{Apolloscapes} dataset \cite{wang2019apolloscape}, featuring 854 training and 400 test images, provides pixel-level labels for 23 classes like vehicles and pedestrians. The \textbf{CARLA\_ADV} dataset, generated via the CARLA simulator (version 0.9.13) \cite{dosovitskiy2017carla}, focuses on adverse weather conditions (e.g., fog, rain) and includes 2,764 training and 1,921 test images, annotated for 23 pixel-level classes such as vehicles and trees.

\subsubsection{Evaluation Metrics}
We evaluate MoE-RAM on AD semantic segmentation task using four metrics: \textbf{mIoU}, which quantifies the overlap between prediction and ground truth; \textbf{mPre}, which measures the accuracy of positive prediction; \textbf{mRec}, which evaluates the model's ability to identify relevant instances; and \textbf{mF1}, which balances mPre and mRec.

\subsubsection{Implementation Details}
For optimization, the Adam optimizer is chosen with Betas values of 0.9 and 0.999, and a weight decay of 1e-4. All experts within MoE-RAM adopt the ASSP architecture \cite{chen2017rethinking} and are trained with a learning rate of 3e-4. The default configurations of the proposed  MoE-RAM include the following hyperparameters: 10 experts, a TopK value of 5, 16 FRL prototypes per expert, a load-balancing weight ($\lambda_{\text{LB}}$) of 0.01, and a FRL regularization weight ($\lambda_{\text{FRL}}$) of 1e-4. 

For evaluation, on the one hand, we compare the proposed MoE-RAM with other MoE routing strategies, such as LinearMoE \cite{fan2022m3vit}, NonlinearMoE \cite{nguyen2024statistical}, and SoftMoE \cite{puigcerver2023sparse}. We should notice that MoE-RAM and these MoE baselines share the same model architecture, where a ViT backbone is used to extract the latent features of inputs, and multiple parallel ASSP experts are integrated into the MoE to serve as the downstream task head. On the other hand, we also compare MoE-RAM with other single-model methods, such as BiSecNetV2 \cite{yu2021bisenet}, SegNet \cite{badrinarayanan2017segnet}, SegFormer \cite{xie2021segformer}, AttaNet \cite{song2021attanet}, HRDA \cite{hoyer2023domain}, TopFormer \cite{zhang2022topformer}, and SeaFormer \cite{wan2023seaformer}.

Notably, the proposed MoE-RAM, all MoE baselines, and all single-model competitors are implemented using the PyTorch framework and trained on two NVIDIA GeForce 4090 GPUs. For MoE-based models, including MoE-RAM, LinearMoE, NonlinearMoE, and SoftMoE, a pretrained ViT backbone is used and frozen, with only the experts (\ie, ASSPs) within the MoE being trained. In contrast, all single-model competitors are trained from scratch using the adopted AD datasets.

\subsection{Main Results and Empirical Analyses} \label{sec:exp_comp_analyses}
In this section, we present experimental comparison of MoE-RAM against other MoE and single-model baselines, including performance and convergence comparisons. 

\begin{table*}[tp]
\setlength{\tabcolsep}{4.3pt}
\centering
\caption{Performance comparison of the proposed MoE-RAM agaist other MoE baselines and single-model approaches for all adopted metrics across multiple AD datasets}
\begin{tabularx}{\linewidth}{c|cccc|cccc|cccc|cccc}
\hline
\multirow{2}{*}{Method} & \multicolumn{4}{c|}{Apolloscapes}                                                                            & \multicolumn{4}{c|}{CamVid}                                                                                  & \multicolumn{4}{c|}{CARLA\_ADV}                                                                              & \multicolumn{4}{c}{Cityscapes}                                                                              \\ \cline{2-17} 
                        & \multicolumn{1}{c}{mIoU}  & \multicolumn{1}{c}{mF1}   & \multicolumn{1}{c}{mPre} & \multicolumn{1}{c|}{mRec} & \multicolumn{1}{c}{mIoU}  & \multicolumn{1}{c}{mF1}   & \multicolumn{1}{c}{mPre} & \multicolumn{1}{c|}{mRec} & \multicolumn{1}{c}{mIoU}  & \multicolumn{1}{c}{mF1}   & \multicolumn{1}{c}{mPre} & \multicolumn{1}{c|}{mRec} & \multicolumn{1}{c}{mIoU}  & \multicolumn{1}{c}{mF1}   & \multicolumn{1}{c}{mPre} & \multicolumn{1}{c}{mRec} \\ \hline
BiSecNetV2              & \multicolumn{1}{c}{22.92} & \multicolumn{1}{c}{27.12} & \multicolumn{1}{c}{-}    & \multicolumn{1}{c|}{-}    & \multicolumn{1}{c}{47.89} & \multicolumn{1}{c}{53.33} & \multicolumn{1}{c}{-}    & \multicolumn{1}{c|}{-}    & \multicolumn{1}{c}{28.80} & \multicolumn{1}{c}{33.59} & \multicolumn{1}{c}{-}    & \multicolumn{1}{c|}{-}    & \multicolumn{1}{c}{33.63} & \multicolumn{1}{c}{43.32} & \multicolumn{1}{c}{-}    & \multicolumn{1}{c}{-}    \\
SegNet                  & \multicolumn{1}{c}{21.01} & \multicolumn{1}{c}{24.60} & \multicolumn{1}{c}{-}    & \multicolumn{1}{c|}{-}    & \multicolumn{1}{c}{46.60} & \multicolumn{1}{c}{50.18} & \multicolumn{1}{c}{-}    & \multicolumn{1}{c|}{-}    & \multicolumn{1}{c}{31.67} & \multicolumn{1}{c}{37.15} & \multicolumn{1}{c}{-}    & \multicolumn{1}{c|}{-}    & \multicolumn{1}{c}{43.13} & \multicolumn{1}{c}{53.87} & \multicolumn{1}{c}{-}    & \multicolumn{1}{c}{-}    \\
SegFormer               & \multicolumn{1}{c}{-}     & \multicolumn{1}{c}{-}     & \multicolumn{1}{c}{-}    & \multicolumn{1}{c|}{-}    & \multicolumn{1}{c}{39.37} & \multicolumn{1}{c}{46.23} & \multicolumn{1}{c}{-}    & \multicolumn{1}{c|}{-}    & \multicolumn{1}{c}{-}     & \multicolumn{1}{c}{-}     & \multicolumn{1}{c}{-}    & \multicolumn{1}{c|}{-}    & \multicolumn{1}{c}{39.37} & \multicolumn{1}{c}{46.23} & \multicolumn{1}{c}{-}    & \multicolumn{1}{c}{-}    \\
AttaNet                 & 20.89                     & 24.85                     & 26.64                    & 25.67                     & 51.12                     & 58.89                     & 58.83                    & 60.96                     & 28.97                     & 34.46                     & 35.63                    & 34.54                     & 22.96                     & 27.28                     & 26.11                    & 30.87                    \\
HRDA                    & 22.19                     & 27.27                     & 34.39                    & 26.94                     & 64.42                     & 75.65                     & 83.66                    & 71.80                     & 34.98                     & 43.10                     & 52.13                    & 40.74                     & 38.89                     & 49.38                     & 64.30                    & 45.80                    \\
TopFormer               & 20.84                     & 24.71                     & 26.37                    & 25.70                     & 48.50                     & 57.02                     & 59.34                    & 57.52                     & 32.32                     & 38.22                     & 41.61                    & 37.20                     & 22.15                     & 26.70                     & 25.48                    & 30.24                    \\
SeaFormer               & 20.58                     & 24.53                     & 25.54                    & 25.28                     & 47.85                     & 56.62                     & 56.22                    & 58.65                     & 28.68                     & 34.24                     & 37.76                    & 33.20                     & 20.51                     & 24.02                     & 23.36                    & 26.85                    \\ \hline
ViT+ASSP                & \multicolumn{1}{c}{17.11} & \multicolumn{1}{c}{21.18} & \multicolumn{1}{c}{-}    & \multicolumn{1}{c|}{-}    & \multicolumn{1}{c}{68.12} & \multicolumn{1}{c}{77.01} & \multicolumn{1}{c}{-}    & \multicolumn{1}{c|}{-}    & \multicolumn{1}{c}{31.64} & \multicolumn{1}{c}{37.58} & \multicolumn{1}{c}{-}    & \multicolumn{1}{c|}{-}    & \multicolumn{1}{c}{26.31} & \multicolumn{1}{c}{30.37} & \multicolumn{1}{c}{-}    & \multicolumn{1}{c}{-}    \\ \hline
LinearMoE               & 21.92                     & 26.76                     & 37.23                    & 26.79                     & 71.07                     & 81.36                     & 84.96                    & 79.74                     & 34.97                     & 42.95                     & 52.48                    & 40.57                     & 39.80                     & 51.53                     & 66.07                    & 47.62                    \\
NonLinearMoE            & 22.32                     & 27.30                     & 37.18                    & 27.03                     & 72.09                     & 82.28                     & \textbf{85.66}                    & 80.40                     & 35.41                     & 43.50                     & 54.08                    & 41.38                     & 41.74                     & 54.07                     & 65.90                    & 50.15                    \\
SoftMoE                 & 22.22                     & 27.11                     & \textbf{39.02}                    & 26.95                     & 71.54                     & 81.71                     & 85.20                    & 79.54                     & 36.21                     & 44.37                     & 53.39                    & 42.14                     & 42.08                     & 54.45                     & 66.93                    & 49.79                    \\ \hline
\textbf{MoE-RAM (Ours)}          & \textbf{24.75}                     & \textbf{28.47}                     & 38.35                    & \textbf{29.48}                     & \textbf{72.43}                     & \textbf{82.89}                     & 85.31                    & \textbf{80.92}                     & \textbf{37.97}                     & \textbf{45.56}                     & \textbf{55.19}                    & \textbf{43.39}                     & \textbf{44.74}                     & \textbf{55.99}                     & \textbf{68.30}                    & \textbf{53.47}                    \\ \hline
\end{tabularx}
\label{tab:perf_comp}
\vspace{-0.3cm}
\end{table*}

\Cref{tab:perf_comp} presents performance comparison of the proposed MoE-RAM against other MoE routing baselines and single-model competitors. The evaluation uses aforementioned four metrics across CamVid, Cityscapes, Apolloscapes, CARLA\_ADV datasets. From \Cref{tab:perf_comp}, we can observe following patterns: (I) The proposed MoE-RAM achieves better performance compared to other baselines for almost all metrics across all datasets, demonstrating its great superiority over other baselines. This suggests that MoE-RAM consistently produces prediction that better match the ground truth. The reason behind it can be attributed to the proposed routing and aggregating mechanisms within MoE-RAM. (II) The limited performance improvement of ViT-based methods (including ViT+ASSP, LinearMoE, NonlinearMoE, SoftMoE, and our MoE-RAM) over single-model architectures can be attributed to the fact that these methods only train the downstream head while keeping the ViT backbone frozen, without fine-tuning it. (III) The performance improvement of MoE methods (including LinearMoE, NonlinearMoE, SoftMoE, and our MoE-RAM) over ViT+ASSP can be attributed to the multiple parallel ASSPs (\ie, experts) in such MoE methods. (IV) The performance improvement of our proposed MoE-RAM over other MoE routing baselines (including LinearMoE, NonlinearMoE, and SoftMoE) can be attributed to our proposed statistic-augmented, decoupled routing strategy (\ie, MoE-RM) and aggregating strategy (\ie, MoE-AM), which suggests their superiority over existing MoE routing strategies.

\begin{figure}[tp]
\vspace{-0.1cm}
\centering
\subfloat[\footnotesize mIoU]{\includegraphics[width=0.5\linewidth, height=0.3\linewidth]{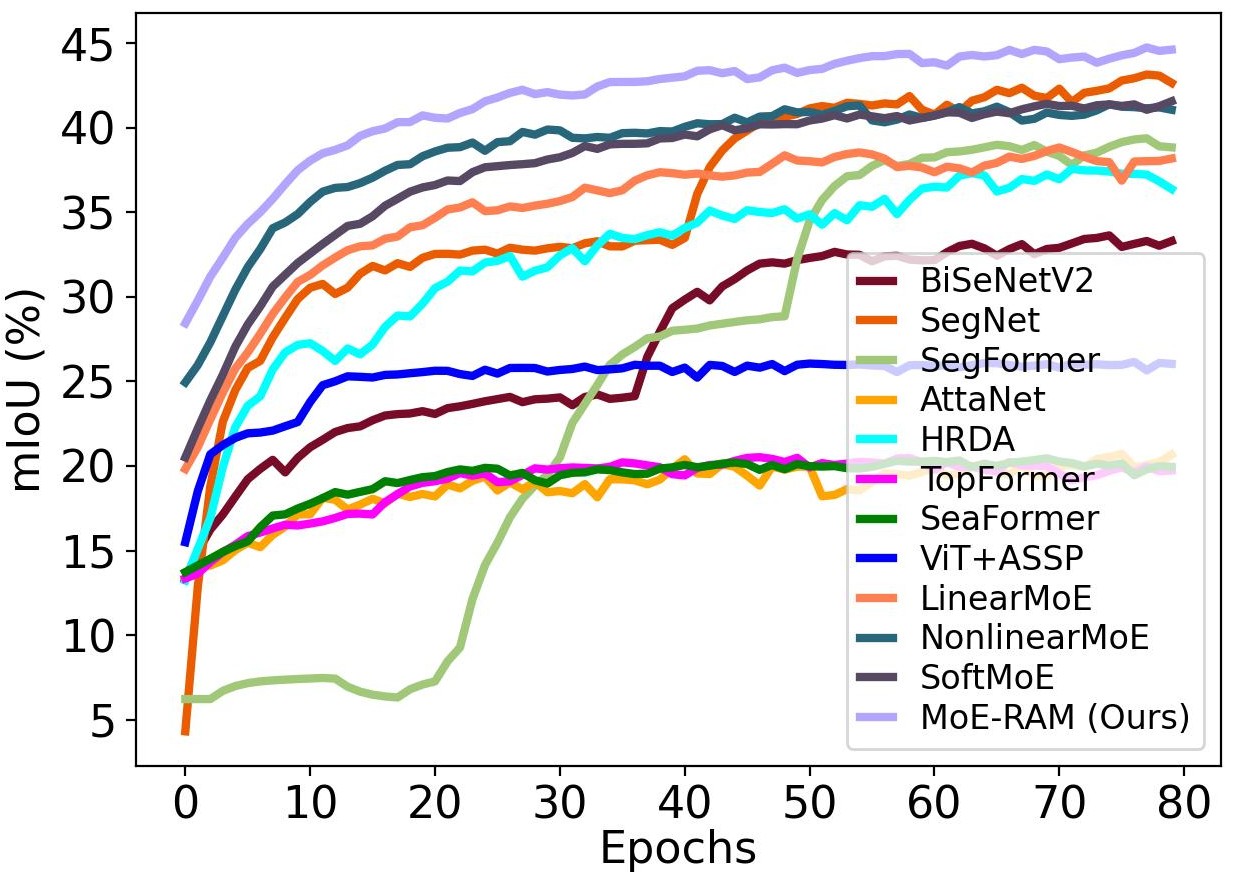}%
\label{Fig:city_mIoU}}
\subfloat[\footnotesize mF1]{\includegraphics[width=0.5\linewidth, height=0.3\linewidth]{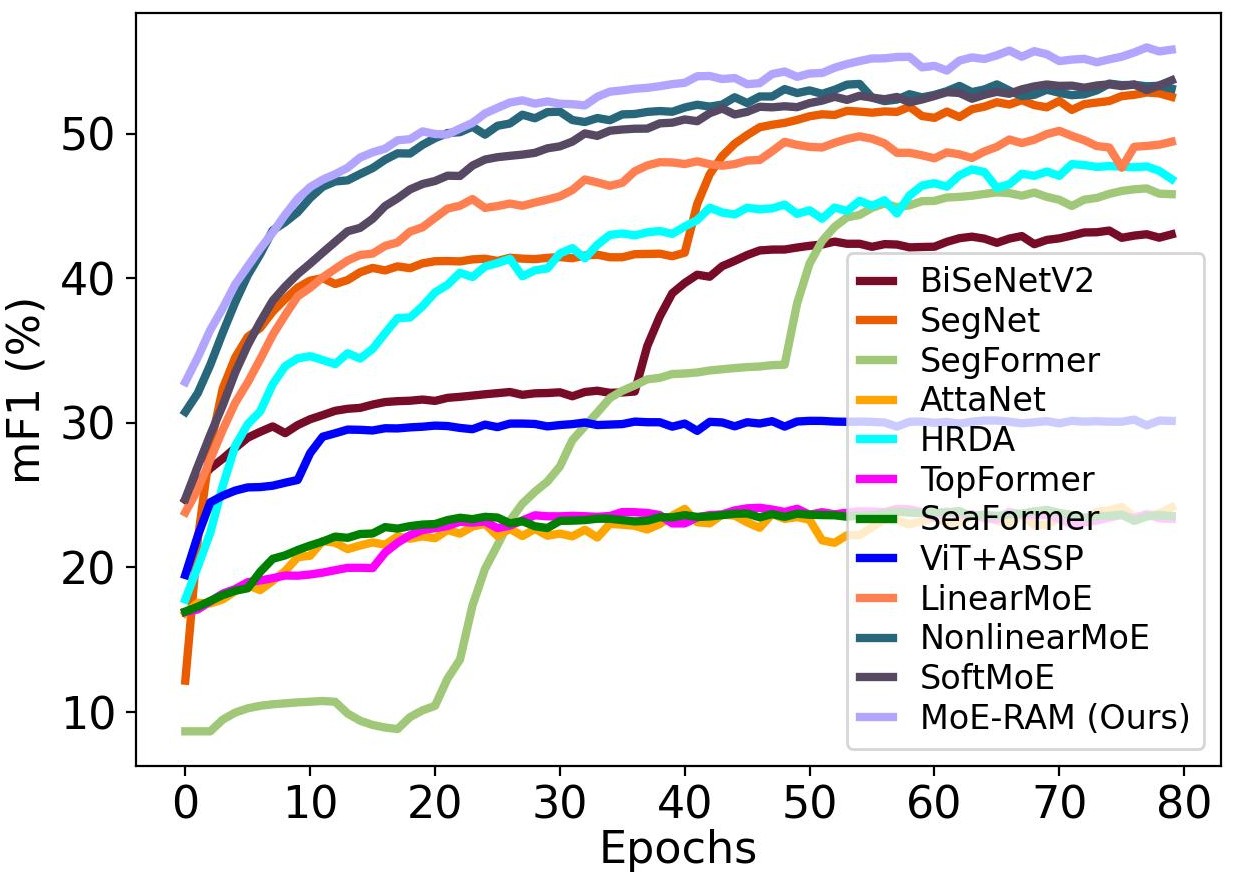}%
\label{Fig:city_mF1}}
\vspace{-0.2cm}
\caption{Convergence comparison of the proposed MoE-RAM against other MoE baselines and single-model approaches.}
\label{Fig:convergece_comp}
\vspace{0.2cm}
\end{figure}

\begin{figure}[tp]
\vspace{-0.2cm}
\centering
\subfloat[\footnotesize T-SNE]{\includegraphics[width=0.5\linewidth]{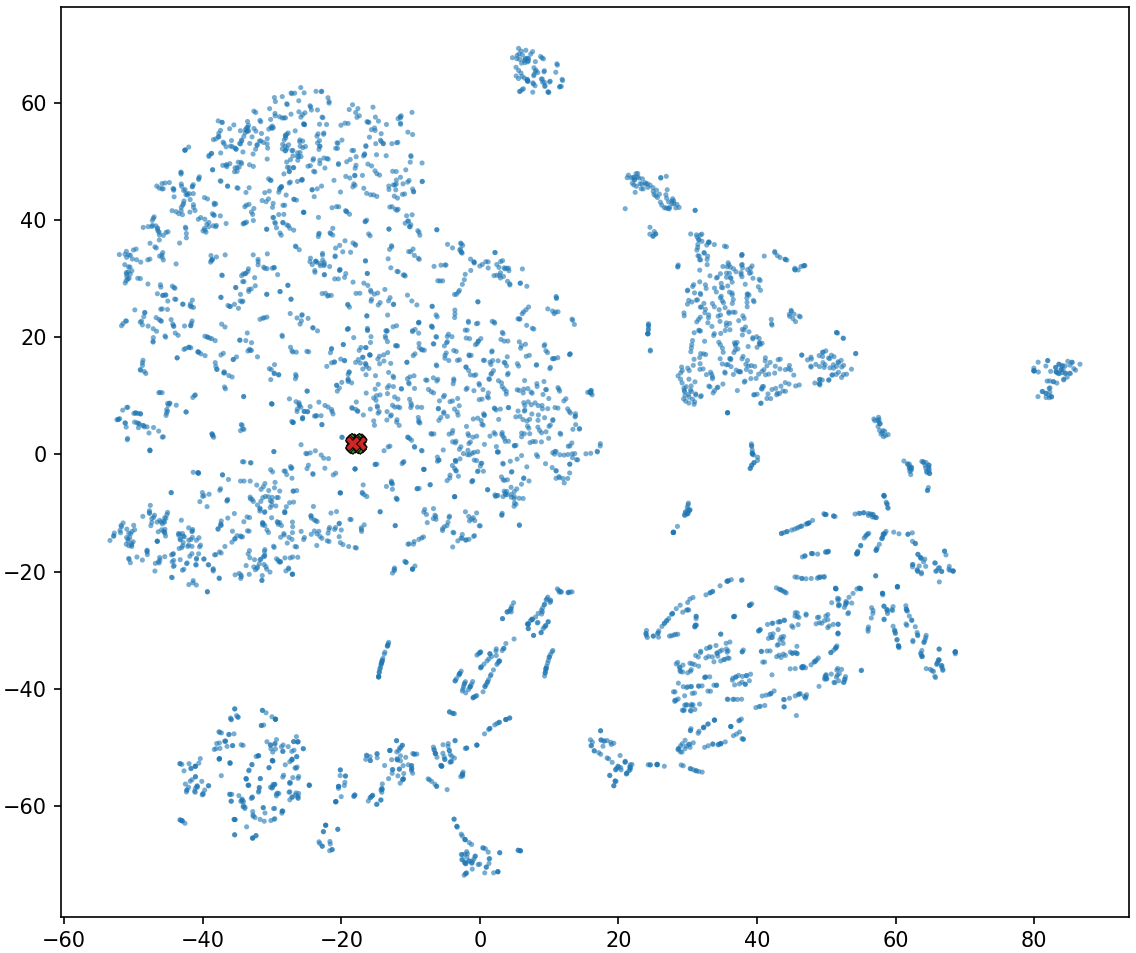}%
\label{Fig:vis_tsne}}
\subfloat[\footnotesize UMAP]{\includegraphics[width=0.5\linewidth]{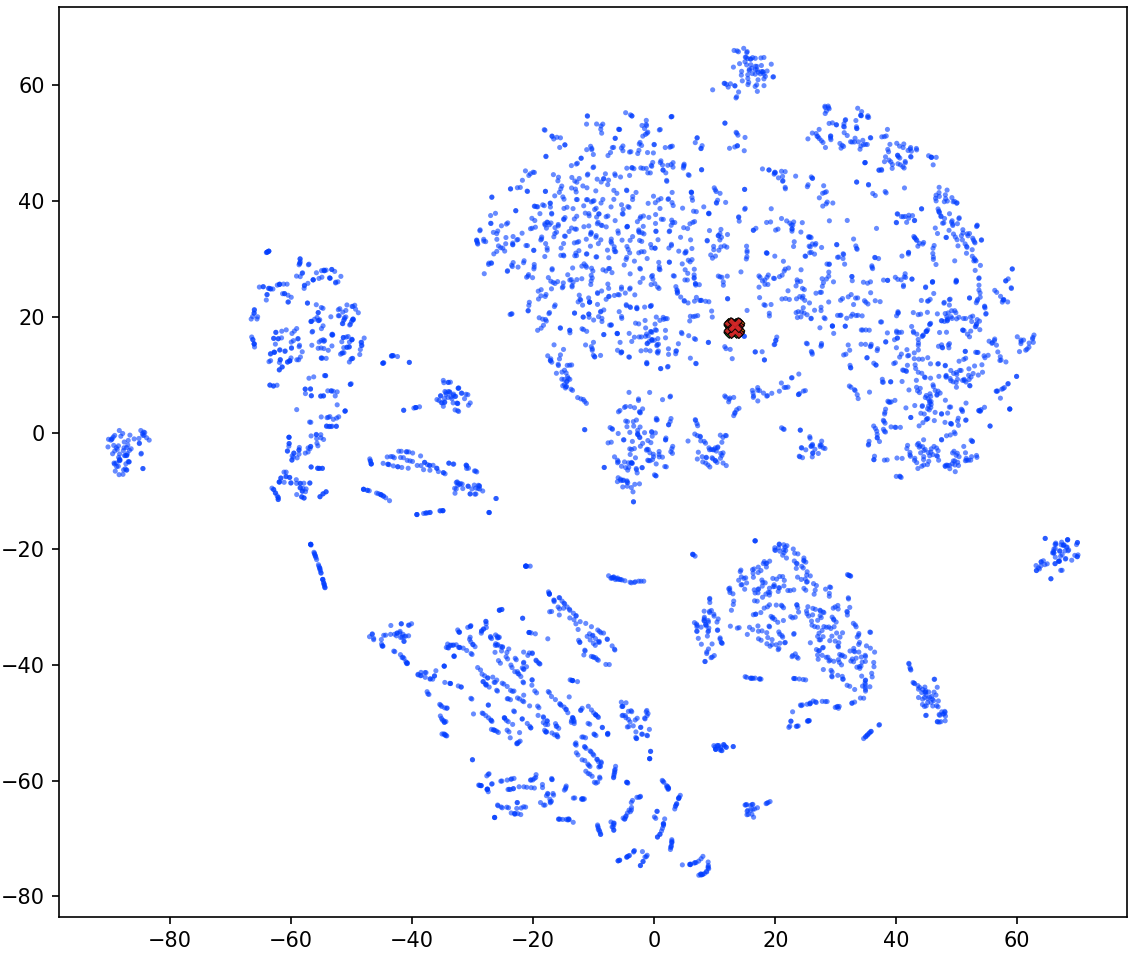}%
\label{Fig:vis_umap}}
\vspace{-0.2cm}
\caption{Relationship visualization between ViT-extracted features and expert-wise FRL prototypes.}
\label{Fig:MoE_RAM_visualization}
\end{figure}

In addition, we also compare the convergence of the proposed MoE-RAM against other MoE routing baselines and single-model competitors, and the results can be viewed in \Cref{Fig:convergece_comp}. From \Cref{Fig:convergece_comp}, we can derive the following insights: (I) ViT backbone-based methods (including ViT+ASSP, LinearMoE, NonlinearMoE, SoftMoE, and our MoE-RAM) generally converge faster than single-model approaches. This can be attributed to the pretraining of ViT on a vast number of datasets. (II) The proposed MoE-RAM converges faster than other MoE routing baselines, thanks to the statistic-augmented, decoupled routing and aggregating strategies.

\subsection{Visualization of the Relationship between Expert-wise FRL Prototypes and ViT-extracted Features} \label{sec:MoERAM_visualization}
This section visualizes the relationship between ViT-extracted feature representations and the most relevant expert's FRL prototypes. The results are presented in \Cref{Fig:MoE_RAM_visualization}, which includes two 2D scatter plots generated using t-SNE (\Cref{Fig:vis_tsne}) and UMAP (\Cref{Fig:vis_umap}), respectively. Each plot overlays sampled ViT token features, represented by smaller points, and FRL prototypes, depicted as larger “X” markers. From \Cref{Fig:MoE_RAM_visualization}, it can be observed that the learned prototypes of the most relevant expert align closely with the majority of the ViT-extracted token features, demonstrating the effectiveness of the proposed expert-wise FRLs.

\begin{table}[tp]
\setlength{\tabcolsep}{3.9pt}
\caption{Performance comparison of cases with different expert numbers within the proposed MoE-RAM}
\begin{tabularx}{\linewidth}{c|c|cccc}
\hline
\multirow{2}{*}{\begin{tabular}[c]{@{}c@{}}Expert\\ Number\end{tabular}} & \multirow{2}{*}{\begin{tabular}[c]{@{}c@{}}MoE-RAM\\ Remaining Configs\end{tabular}}                                                                                     & \multicolumn{4}{c}{Cityscapes} \\ \cline{3-6} 
                                                                         &                                                                                                                                                                          & mIoU   & mF1   & mPre  & mRec  \\ \hline
7                                                                        & \multirow{4}{*}{\begin{tabular}[c]{@{}c@{}}TopK =5,\\ FRL Prototype Number = 16,\\ $\lambda_{\text{LB}}$ = 0.01,\\ $\lambda_{\text{FRL}}$ = 1e-4\end{tabular}} & \textbf{45.45}  & \textbf{56.76} & \textbf{71.65} & 53.22 \\
8                                                                        &                                                                                                                                                                          & 45.02  & 56.08 & 70.19 & 53.68 \\
9                                                                        &                                                                                                                                                                          & 44.99  & 55.82 & 69.07 & \textbf{53.87} \\
10                                                                       &  & 44.74                     & 55.99                     & 68.30                    & 53.47  \\ \hline 
\end{tabularx}
\label{tab:abl_expert_num}
\end{table}

\subsection{Ablation Study} \label{sec:ablation_study}
This section investigates how MoE-RAM-embedded hyperparameters (including the number of experts, the TopK value, the FRL prototype number, $\lambda_{\text{LB}}$, and $\lambda_{\text{FRL}}$) affect the overall performance of MoE-RAM.

\subsubsection{The Influence of the Number of Experts on the Overall Performance of MoE-RAM}
\Cref{tab:abl_expert_num} showcases the influence of the number of experts on the overall performance of MoE-RAM. From this table, we can find that the case with 7 experts achieves the best performance across almost all metrics. This inspires us that, when the TopK value is fixed, as the number of experts increases, the performance of MoE-RAM declares gradually. In other words,  the performance of MoE-RAM is positively related to the proportion of TopK value to the number of experts included in MoE-RAM.

\begin{table}[tp]
\setlength{\tabcolsep}{4.6pt}
\caption{Performance comparison of cases with different TopK values in expert routing of MoE-RAM}
\begin{tabularx}{\linewidth}{c|c|cccc}
\hline
\multirow{2}{*}{\begin{tabular}[c]{@{}c@{}}TopK\\ Value\end{tabular}} & \multirow{2}{*}{\begin{tabular}[c]{@{}c@{}}MoE-RAM\\ Remaining Configs\end{tabular}}                                                                                               & \multicolumn{4}{c}{Cityscapes}                                                                           \\ \cline{3-6} 
                                                                      &                                                                                                                                                                                    & \multicolumn{1}{c}{mIoU} & \multicolumn{1}{c}{mF1} & \multicolumn{1}{c}{mPre} & \multicolumn{1}{c}{mRec} \\ \hline
3                                                                     & \multirow{4}{*}{\begin{tabular}[c]{@{}c@{}}Expert Number =10,\\ FRL Prototype Number = 16,\\ $\lambda_{\text{LB}}$ = 0.01,\\ $\lambda_{\text{FRL}}$ = 1e-4\end{tabular}} & \textbf{45.85}                    & \textbf{56.75}                   & \textbf{69.91}                    & \textbf{53.69}                    \\
4                                                                     &                                                                                                                                                                                    & 45.46                    & 55.47                   & 68.82                    & 53.41                    \\
5                                                                     &  & 44.74                     & 55.99                     & 68.30                    & 53.47                                                                                                                                                                                                            \\
6                                                                     &                                                                                                                                                                                    & 44.33                    & 55.65                   & 67.44                    & 52.48                    \\ \hline
\end{tabularx}
\label{tab:abl_topk_value}
\end{table}

\begin{table}[tp]
\setlength{\tabcolsep}{4.6pt}
\caption{Performance comparison of cases with different FRL prototype numbers in the proposed MoE-RAM}
\begin{tabularx}{\linewidth}{c|c|cccc}
\hline
\multirow{2}{*}{\begin{tabular}[c]{@{}c@{}}FRL Prototype\\ Number\end{tabular}} & \multirow{2}{*}{\begin{tabular}[c]{@{}c@{}}MoE-RAM\\ Remaining Configs\end{tabular}}                                                                              & \multicolumn{4}{c}{Cityscapes} \\ \cline{3-6} 
                                                                                &                                                                                                                                                                   & mIoU   & mF1   & mPre  & mRec  \\ \hline
4                                                                               & \multirow{4}{*}{\begin{tabular}[c]{@{}c@{}}Expert Number =10,\\ TopK = 5,\\ $\lambda_{\text{LB}}$ = 0.01,\\ $\lambda_{\text{FRL}}$ = 1e-4\end{tabular}} & \textbf{45.59}  & \textbf{57.35} & \textbf{70.76} & 53.84 \\
8                                                                               &                                                                                                                                                                   & 45.27  & 56.92 & 69.16 & 53.68 \\
12                                                                              &                                                                                                                                                                   & 44.47  & 55.25 & 67.99 & \textbf{54.04} \\
16                                                                              &                                                                                                                                                                   & 44.74                     & 55.99                     & 68.30                    & 53.47       \\ \hline
\end{tabularx}
\label{tab:abl_FRL_proto_num}
\end{table}

\subsubsection{The Effect of the TopK Value on the Overall Performance of MoE-RAM}
\Cref{tab:abl_topk_value} illustrates the effect of the TopK value on the overall performance of MoE-RAM. We can observe from the table that the case with TopK value being 6 performs better than other cases with TopK values ranging from 3 to 5. This shows that when the number of experts within MoE-RAM is fixed, the overall performance of MoE-RAM increases with respect to the TopK value. This further demonstrates that the performance of MoE-RAM is positively correlated with the ratio of the TopK value to the total number of experts included in the proposed MoE-RAM.

\subsubsection{The Impact of the FRL Prototype Number on the Overall Performance of MoE-RAM}
\Cref{tab:abl_FRL_proto_num} demonstrates the impact of the number of experts' FRL prototype for the overall performance of MoE-RAM. From this table, we can observe that the case with an FRL prototype number of 4 outperforms the cases with FRL prototype numbers of 8, 12, and 16, respectively. This suggests that MoE-RAM's overall performance is negatively impacted by an increase in the number of FRL prototypes. Therefore, in practical implementations, it is advisable to set a relatively small number of prototypes in the expert-wise FRL.

\begin{table}[tp]
\setlength{\tabcolsep}{5.0pt}
\caption{Performance comparison of cases with different $\lambda_{\text{LB}}$ values in the proposed MoE-RAM}
\begin{tabularx}{\linewidth}{c|c|llll}
\hline
\multirow{2}{*}{$\lambda_{\text{LB}}$} & \multirow{2}{*}{\begin{tabular}[c]{@{}c@{}}MoE-RAM\\ Remaining Configs\end{tabular}}                                                                      & \multicolumn{4}{c}{Cityscapes}                                                                           \\ \cline{3-6} 
                           &                                                                                                                                                           & \multicolumn{1}{c}{mIoU} & \multicolumn{1}{c}{mF1} & \multicolumn{1}{c}{mPre} & \multicolumn{1}{c}{mRec} \\ \hline
0.00                       & \multirow{5}{*}{\begin{tabular}[c]{@{}c@{}}Expert Number =10,\\ TopK = 5,\\ FRL Prototype Number = 16,\\ $\lambda_{\text{FRL}}$ = 1e-4\end{tabular}} & 44.23                    & 55.48                   & \textbf{71.13}                    & 50.82                    \\
0.01                       &     & 44.74                     & 55.99                     & 68.30                    & \textbf{53.47}                          \\
0.02                       &                                                                                                                                                           & 43.97                    & 55.19                   & 68.87                    & 52.06                    \\
0.04                       &                                                                                                                                                           & \textbf{45.44}                    & \textbf{57.17}                   & 70.86                    & 53.09                    \\
0.08                       &                                                                                                                                                           & 44.04                    & 55.46                   & 69.30                    & 51.78                    \\ \hline
\end{tabularx}
\label{tab:abl_LB_weight}
\end{table}

\begin{table}[!tp]
\setlength{\tabcolsep}{4.8pt}
\caption{Performance comparison of cases with different $\lambda_{\text{FRL}}$ values in the proposed MoE-RAM}
\begin{tabularx}{\linewidth}{c|c|cccc}
\hline
\multirow{2}{*}{$\lambda_{\text{FRL}}$} & \multirow{2}{*}{\begin{tabular}[c]{@{}c@{}}MoE-RAM\\ Remaining Configs\end{tabular}}                                                                     & \multicolumn{4}{c}{Cityscapes}                                                                           \\ \cline{3-6} 
                            &                                                                                                                                                          & \multicolumn{1}{c}{mIoU} & \multicolumn{1}{c}{mF1} & \multicolumn{1}{c}{mPre} & \multicolumn{1}{c}{mRec} \\ \hline
0.00                        & \multirow{5}{*}{\begin{tabular}[c]{@{}c@{}}Expert Number =10,\\ TopK = 5,\\ FRL Prototype Number = 16,\\ $\lambda_{\text{LB}}$ = 0.01\end{tabular}} & 44.23                    & 55.71                   & 68.62                    & 52.18                    \\
1e-4                        &   & 44.74                     & 55.99                     & 68.30                    & 53.47                          \\
2e-4                        &                                                                                                                                                          & 44.16                    & 55.47                   & 69.65                    & 52.19                    \\
4e-4                        &                                                                                                                                                          & \textbf{45.44}                    & \textbf{57.17}                   & \textbf{71.17}                    & \textbf{53.60}                    \\
8e-4                        &                                                                                                                                                          & 45.03                    & 56.44                   & 69.93                    & 52.29                    \\ \hline
\end{tabularx}
\label{tab:abl_FRL_weight}
\end{table}

\subsubsection{The Effect of $\lambda_{\text{LB}}$ on the Overall Performance of MoE-RAM}
\Cref{tab:abl_LB_weight} indicates how $\lambda_{\text{LB}}$ imposes the influence on the overall performance of MoE-RAM. We can conclude from this table that the case $\lambda_{\text{LB}}=0.04$ outperforms the case $\lambda_{\text{LB}}=0.00$, indicating that selecting an appropriate value of $\lambda_{\text{LB}}$ can improve the overall performance of MoE-RAM. However, the cases $\lambda_{\text{LB}}=0.02$ and $\lambda_{\text{LB}}=0.08$ perform worse than the case $\lambda_{\text{LB}}=0.00$. This suggests that inappropriate setting $\lambda_{\text{LB}}$ can reduce the overall performance of MoE-RAM. In conclusion, proper tuning of $\lambda_{\text{LB}}$ is essential for optimizing the performance of MoE-RAM in practical implementations.

\subsubsection{The Impact of $\lambda_{\text{FRL}}$ on the Overall Performance of MoE-RAM}
\Cref{tab:abl_FRL_weight} compares the results of different $\lambda_{\text{FRL}}$ values on the overall performance of MoE-RAM. We can observe from \Cref{tab:abl_FRL_weight} that the settings $\lambda_{\text{FRL}}=1e-4$, $\lambda_{\text{FRL}}=4e-4$, and $\lambda_{\text{FRL}}=8e-4$ outperform the setting $\lambda_{\text{FRL}}=0.00$, indicating that selecting an appropriate value of $\lambda_{\text{FRL}}$ can enhance the overall performance of MoE-RAM. However, the case $\lambda_{\text{FRL}}=2e-4$ performs worse than the case $\lambda_{\text{FRL}}=0.00$, suggesting that an inappropriate setting of $\lambda_{\text{FRL}}$ can negatively impact the overall performance of MoE-RAM. In conclusion, careful tuning of $\lambda_{\text{FRL}}$ is crucial for optimizing the performance of MoE-RAM in practical implementations.

\section{Conclusion}
AD scenarios are inherently complex and diverse, presenting significant challenges for a single model to handle varying conditions such as weather, traffic density, and road types. LM-Driven MoE offers a promising solution, where the LM acts as the backbone to extract latent features, while the MoE serves as the downstream head, dynamically selecting and aggregating specialized experts to adapt to different scenarios. However, MoE faces intrinsic challenges like flawed routing strategies and inefficient expert aggregation. To address these limitations, the proposed MoE-RAM incorporates MoE-RM to select the most relevant experts, and integrates MoE-AM to reweight experts' outputs during fusion. Taking AD semantic segmentation as an example task, extensive experiments on AD datasets demonstrate the superiority of MoE-RAM compared to other MoE routing baselines and conventional single-model methods.

\end{document}